\documentclass[10pt, a4paper]{article}

\usepackage[final]{lrec2026} 
\usepackage{amsmath}
\usepackage{amssymb} 
\usepackage{tabularx} 
\usepackage{graphicx}
\usepackage{csquotes}

\title{Semantic Centroids and Hierarchical Density-Based Clustering for Cross-Document Software Coreference Resolution}

\name{Julia Matela, Frank Krüger} 

\address{Department of Electrical Engineering and Computer Science, Wismar University of Applied Sciences\\
         Wismar, Germany\\
         julia.matela@hs-wismar.de\\
        }

\abstract{
This paper describes the system submitted to the SOMD 2026 Shared Task for Cross-Document Coreference Resolution (CDCR) of software mentions. 
Our approach addresses the challenge of identifying and clustering inconsistent software mentions across scientific corpora. 
We propose a hybrid framework that combines dense semantic embeddings from a pre-trained Sentence-BERT model, Knowledge Base (KB) lookup strategy built from training-set cluster centroids using FAISS for efficient retrieval, and HDBSCAN density-based clustering for mentions that cannot be confidently assigned to existing clusters. 
Surface-form normalization and abbreviation resolution are applied to improve canonical name matching. 
The same core pipeline is applied to Subtasks 1 and 2.
To address the large scale settings of Subtask 3, the pipeline was adapted by utilising a blocking strategy based on entity types and canonicalized surface forms. 
Our system achieved CoNLL F1 scores of 0.98, 0.98, and 0.96 on Subtasks 1, 2, and 3 respectively.
\\ \newline  }

\begin{document}

\maketitleabstract

\section{Task and Data}

The SOMD 2026 Shared Task addresses an issue of cross-document coreference resolution (CDCR).
The task is structured into three subtasks.
For each subtask participants are provided with a set of software mentions that were extracted from \textsc{SoMeSci}~\cite{Schindler2021} and SoftwareKG~\cite{Schindler_2022}.
The objective is to partition them into clusters such that each cluster contains mentions that refer to the same underlying software. 

The dataset for Subtask 1 consists of gold-standard annotated mentions.
The training set contains 2974 mentions partitioned into 733 clusters and the test set contains 743 mentions.
Subtask 2 and 3 both use automatically predicted mentions from an upstream extraction pipeline, introducing potential noise. 
The training set contains 2860 mentions assigned in 699 clusters and is identical for Subtask 2 and 3.
The test set for Subtask 2 contains 12516 mentions.
Subtask 3 creates additional computational challenges with a 219950-mention test set.

\section{System Description}

\begin{figure}[!ht]
\begin{center}
\includegraphics[width=\columnwidth]{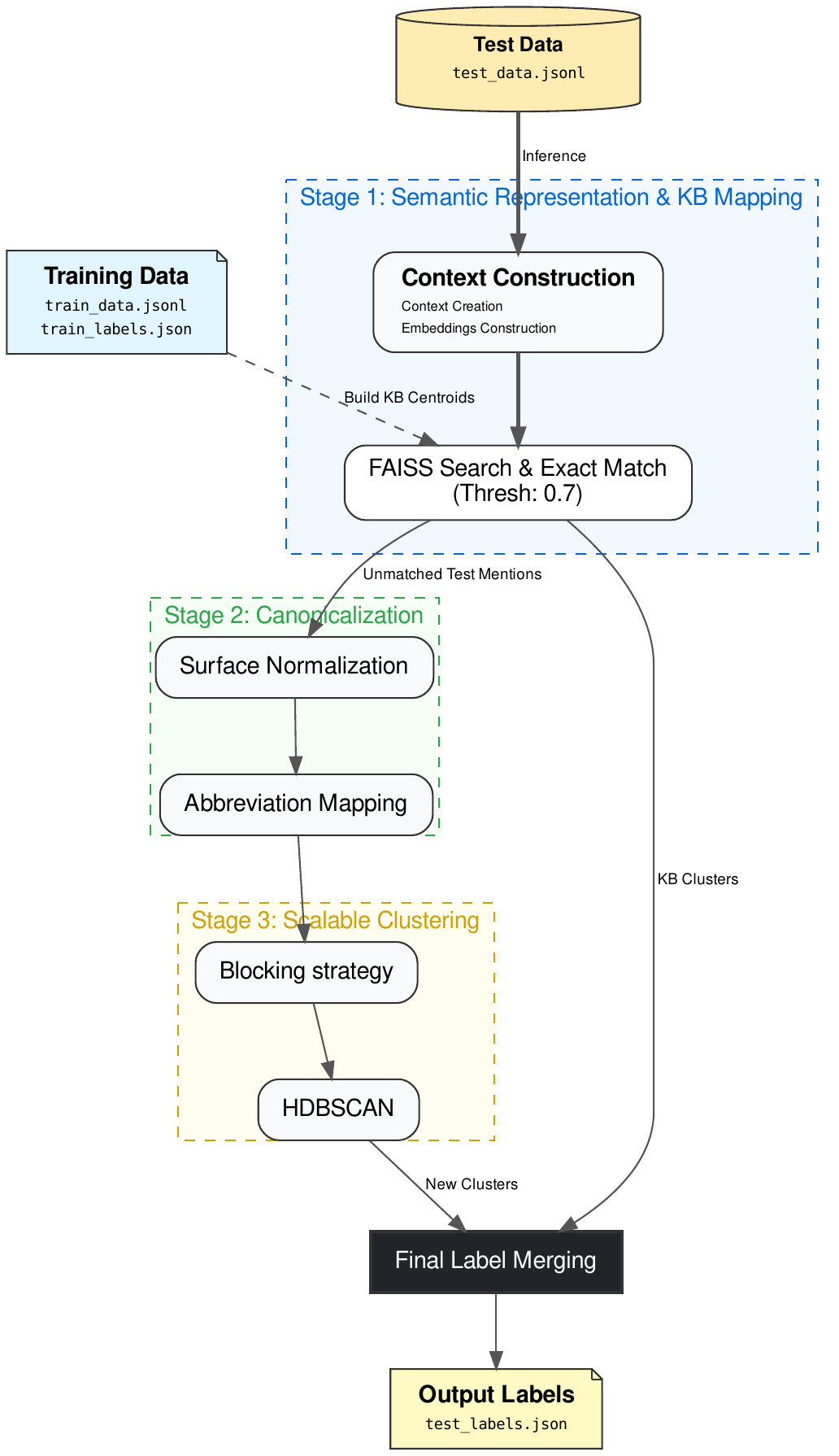}
\caption{System Description}
\label{fig.1}

\end{center}

\end{figure}

Our pipeline follows a three-stage architecture: 
(1) semantic representation and KB centroid construction from training data, 
(2) test-mention assignment to existing KB clusters via FAISS \citep{johnson_billion-scale_2021} retrieval and exact string matching, and 
(3) density-based clustering of unmatched mentions via HDBSCAN.
Figure \ref{fig.1} illustrates the complete pipeline.


\subsection{Semantic Representation}

The semantic representation stage of our pipeline transforms each software mention into
a normalized 384-dimensional embedding that captures both the surface-form identity and
the structured context of the mention.

\subsubsection{Context String Construction}

Software entities in scientific text are highly ambiguous. 
The same tool may be referred to by its full name (\emph{Statistical Package for the Social Sciences}),
an acronym (\emph{SPSS}), or a version-qualified variant (\emph{SPSS 28}).

The surface form itself is therefore insufficient for the task of cross-document coreference resolution.
For that reason, we construct a short, structured context string $c(m)$ for each mention $m$:

\begin{equation}
  c(m) = \underbrace{\text{mention} \parallel \text{mention}}_{\text{weighted anchor}}
         \parallel \text{type}
         \parallel \underbrace{\text{rel}_1 \parallel \text{rel}_2 \parallel \cdots}_{\text{relational context}}
\end{equation}

\noindent where $\parallel$ denotes whitespace-separated concatenation. 
The string is built from three components. 
First, the mention surface form is lowercased and duplicated. 
Second, the entity type label (e.g., \emph{application}, 
\emph{plugin}) is appended in its original casing. Third, the surface forms of
\emph{all} relational mentions present in the metadata are appended.

\paragraph{}
Since the software name is the primary coreference signal, 
the embedding must be anchored to it rather than to surrounding metadata.

Mean-pooling distributes representational weight evenly across all input tokens, 
so a long metadata-heavy string risks diluting the name's contribution and causing so called \emph{semantic drift}. 
We counteract this by duplicating the surface form mention token \cite{Wu2021}, 
effectively increasing its weight in the final embedding.

\subsubsection{Sentence Embedding with \texttt{all-MiniLM-L6-v2}}

Context strings are encoded with the \texttt{all-MiniLM-L6-v2} Sentence-BERT model \citep{reimers-2019}. 

The \texttt{all-MiniLM-L6-v2} checkpoint combines this fine-tuning strategy with the MiniLM compression method \citep{wang_minilm_2020}, 
which distills a 12-layer BERT teacher into a compact 6-layer, 384-dimensional student by mimicking the teacher's self-attention distributions in the final layer.

The result is a model of approximately 22~million parameters, roughly five times smaller than BERT-base, that retains competitive semantic similarity performance while encoding around 14,000 sentences per second on a standard CPU as stated by the developers \citep{muennighoff_mteb_2023}. 
This throughput is decisive for Subtask~3, where the mention inventory is an order of magnitude larger than in Subtasks~1 and~2.

\paragraph{Pooling and normalization.}
Token representations from the final Transformer layer are aggregated by attention-mask-weighted mean pooling:

\begin{equation}
  \mathbf{e}(m) = \frac{\sum_{i} a_i \mathbf{h}_i}{\sum_{i} a_i}
\end{equation}

where $\mathbf{h}_i \in \mathbb{R}^{384}$ is the $i$-th token representation and $a_i \in \{0,1\}$ is its attention mask value. 
The pooled vector is then $L_2$-normalized to unit length, converting cosine similarity into an inner product and enabling efficient search via the FAISS \texttt{IndexFlatIP} index \citep{johnson_billion-scale_2021}.

\subsection{Knowledge Base Centroid Construction}

For the training set, we compute a single $L_2$-normalized centroid vector for each gold cluster by averaging the embeddings of all member mentions and re-normalizing.
These centroids form a Knowledge Base (KB) of known software identities.

During inference, test mentions are queried against these centroids using FAISS for efficient Inner Product (cosine) similarity search.

\subsection{Test-Mention Assignment}
Each test mention is either assigned to a KB cluster, or flagged as unmatched following the procedure:

\paragraph{\textbf{1. Exact string match} (Subtask 3).}

If the lowercase and stripped canonical name of a test mention is found within the KB string dictionary derived from training data, 
the corresponding cluster is assigned.

\paragraph{\textbf{2. High-confidence semantic match.} }
If the cosine similarity to the closest KB centroid exceeds a threshold of $\theta = 0.7$, 
the mention is allocated to that cluster. 
This threshold was determined through empirical evaluation of various values on the training data.

\paragraph{\textbf{3. Medium-confidence string-corroborated match} (Subtasks 1 and 2) }
If the similarity is lower than a threshold of 0.7, but higher than 0.5 
and the mention's canonical name is present in the KB string dictionary, 
the existing cluster is prioritized. 

This handles mentions where a canonical name match is present, 
but the embedding similarity is reduced due to too much contextual noise in the embedding.

\paragraph{\textbf{4. Flagging as unmatched.} }
All mentions that do not meet the criteria in the preceding steps are passed to the density clustering stage.

\subsection{Normalization and Abbreviation Handling}

To handle variations like \texttt{SPSS} vs \texttt{Statistical Package for Social Sciences}, we implement a canonicalization step. 
We extract \texttt{Abbreviation} relations from the metadata to build a mapping from short-form to long-form names. 
Surface forms are then normalized by lowercasing and removing non-alphanumeric characters.

\subsection{Clustering Unmatched Mentions}
Mentions that fail to match the KB are processed via \texttt{HDBSCAN} \cite{campello_2013}, 
a density-based algorithm that does not require a pre-specified number of clusters and handles outliers by assigning them to singleton clusters.

To ensure scalability in Subtask 3, we utilize a blocking strategy before clustering.
Mentions are first partitioned by the entity type.
For blocks exceeding 20,000 mentions, we further partition by the first letter of the canonical name.
This reduces the complexity of distance-matrix computation from $O(n^2)$ over the full unmatched set to $O\left(\sum_k n_k^2\right)$ over smaller independent blocks.

Within each block, HDBSCAN is employed with \texttt{min\_cluster\_size=2}, \texttt{min\_samples=1}, and \texttt{cluster\_selection\_epsilon=0.5} parameters for Subtask 1 and 2.
In Subtask 3 the more restrictive epsilon value of 0.15 is implemented to prevent excessive merging within the denser vector space, which arises from much larger collection of mentions.

Following the clustering process, mentions that share a canonical name are merged into a single cluster.
This consolidation occurs regardless of the HDBSCAN's clustering outcome, utilizing the information provided by the canonicalization process.

\subsection{Final Label Merging}
Mentions identified through KB matching and those newly clustered are combined into a unified list of clusters.
The established name of an entity serves as an extra indicator for merging
if a mention's canonical name is already present in a KB cluster, it is merged rather than forming a new cluster.

\section{Experimental Setup}
The system was implemented in Python using \texttt{sentence-transformers} for embeddings, \texttt{faiss-cpu} for approximate nearest-neighbor search, 
and \texttt{scikit-learn} (Subtask 3) or the \texttt{hdbscan} library (Subtasks 1 and 2) for density-based clustering.
All experiments were conducted on a compute system equipped with dual Intel\,Xeon\,Gold\,6346 processors, an NVIDIA\,A100 GPU featuring 80\,GB of VRAM, and 1\,TB of system memory.
The full pipeline code is publicly available at \url{https://github.com/matjulia/somd2026}.

\section{Results}

Our system demonstrated robust performance across all levels of difficulty (Table 1).

\begin{table}[ht]
\centering
\begin{tabularx}{\columnwidth}{|l|X|X|X|}
\hline
\textbf{Metric} & \textbf{Task 1} & \textbf{Task 2} & \textbf{Task 3} \\ \hline
MUC F1 & 0.9939 & 0.9916 & 0.9912 \\ \hline
BCUB F1 & 0.9905 & 0.9858 & 0.9724 \\ \hline
CEAFE F1 & 0.9584 & 0.9521 & 0.9218 \\ \hline
\textbf{CoNLL F1} & \textbf{0.9809} & \textbf{0.9765} & \textbf{0.9618} \\ \hline
\end{tabularx}
\caption{System performance across SOMD 2026 Subtasks.}
\end{table}

Performance is highest on Subtask 1, where gold-standard mentions provide clean surface forms and reliable metadata.
The slight decrease in performance on Subtask 2 reflects the noise introduced by mention extraction/ prediction.
Subtask 3 exhibits a more significant performance decline, attributable to the expanded search space.

\subsection{Scalability Analysis}

\begin{table*}[ht]
\centering
\small
\begin{tabular}{lrrrrrrr}
\hline
\textbf{Fraction} & \textbf{\textit{n}} & \textbf{Embed (s)} & \textbf{KB Match (s)} & \textbf{Canon. (s)} & \textbf{HDBSCAN (s)} & \textbf{Merge (s)} & \textbf{Total (s)} \\
\hline
10\%              & 21{,}995  & $2.23 \pm 0.30$ & $0.60 \pm 0.02$ & $0.09 \pm 0.02$ & $14.02 \pm 0.18$ & $0.12 \pm 0.01$ & $17.06 \pm 0.40$ \\
25\%              & 54{,}987  & $5.54 \pm 0.38$ & $1.51 \pm 0.01$ & $0.33 \pm 0.31$ & $85.94 \pm 1.03$ & $0.30 \pm 0.01$ & $93.62 \pm 1.15$ \\
50\%              & 109{,}975 & $10.97 \pm 0.32$ & $3.02 \pm 0.01$ & $0.73 \pm 0.32$ & $64.97 \pm 0.64$ & $0.58 \pm 0.02$ & $80.27 \pm 0.54$ \\
75\%              & 164{,}962 & $16.87 \pm 0.27$ & $4.47 \pm 0.03$ & $0.68 \pm 0.24$ & $142.29 \pm 0.77$ & $0.90 \pm 0.01$ & $165.20 \pm 0.85$ \\
100\%    & 219{,}950 & $22.32$          & $5.89$          & $0.80$          & $99.01$           & $1.16$          & $129.17$ \\
\hline
\end{tabular}
\caption{ Mean wall-clock time per pipeline stage at each test-set fraction.
         Values for 10\%--75\% are means $\pm$ std over six random seeds;
         the 100\% row is the mean of two independent full runs.}
\label{tab:scalability}
\end{table*}

\begin{figure*}[ht]
\centering
\includegraphics[width=0.76\textwidth]{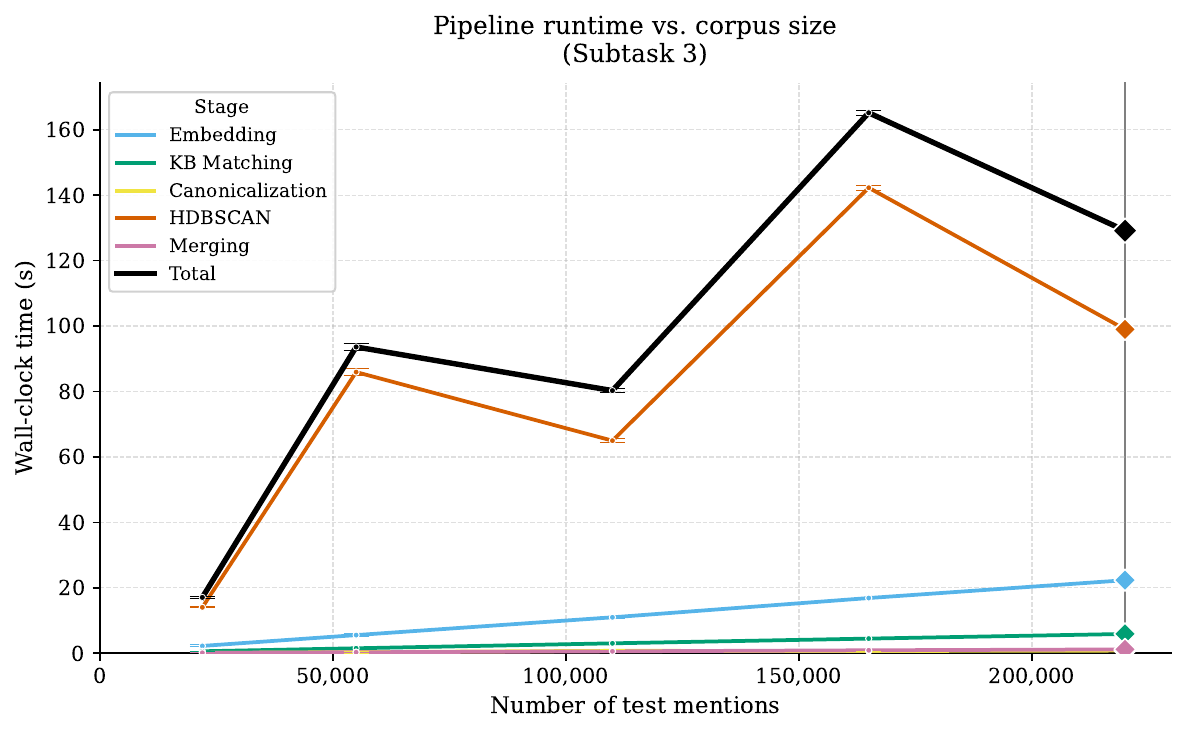}
\caption{Pipeline runtime vs.\ corpus size for Subtask~3.
         Each coloured line shows the mean wall-clock time of one pipeline stage
         across six random test set fractions.
        The bold black line shows total runtime.}
\label{fig:scalability}
\vspace{-0.1cm}
\end{figure*}

To quantify the computational behaviour of the pipeline under increasing data volume, 
we sampled the Subtask 3 test set at fractions of 10\%, 25\%, 50\%, and 75\%, 
evaluating each fraction with six different random seeds (42, 123, 7, 13, 111 and 23).
The full test set (100\%) was run once without sampling.

Wall-clock time was measured independently for each pipeline stage:
(\textit{i})~sentence embedding,
(\textit{ii})~FAISS KB matching,
(\textit{iii})~surface-form canonicalization,
(\textit{iv})~HDBSCAN blocking and clustering, and
(\textit{v})~final label merging.
Results are reported in Table~\ref{tab:scalability} and visualised in Figure~\ref{fig:scalability}.

The embedding and FAISS retrieval stages scale approximately linearly with the
number of mentions, as both operations process each mention independently.
The canonicalization and merging stages are dominated by dictionary lookups and
remain negligible across all corpus sizes.

The HDBSCAN clustering stage exhibits non-monotonic runtime behavior, 
characterized by inflection points in its scaling curve. As shown in 
Table~\ref{tab:scalability}, the 25\% sampling fraction requires a higher 
wall-clock time than the 50\% fraction across all experimental seeds.

This stagnation in runtime is further observed between the 75\% and 100\% 
fractions. This behavior aligns with well-documented challenges in record 
linkage, where a single oversized block can dominate the overall execution 
time~\cite{papadakis_blocking_2021}, and is a direct consequence of the 
hierarchical blocking strategy implemented to mitigate the $O(N^2)$ or 
$O(N \log N)$ scalability bottlenecks inherent to density-based 
clustering~\cite{mcinnes_accelerated_2017}.

\section{Conclusion}

We presented a hybrid pipeline for cross-document coreference resolution of software
mentions, combining \texttt{all-MiniLM-L6-v2} sentence embeddings, Knowledge Base
centroid retrieval via FAISS, surface-form canonicalization, and HDBSCAN clustering.
The system achieves CoNLL F1 scores of 0.98, 0.98, and 0.96 on Subtasks 1, 2, and 3
respectively, with the blocking strategy enabling scalability to corpora exceeding
200{,}000 mentions without GPU acceleration.

We further observe that the task, while framed as coreference resolution, more closely
resembles entity disambiguation: a software may be referred to by multiple surface forms,
but each surface form identifies at most one software.
This asymmetry distinguishes the problem from classical coreference and may inform
more targeted modelling approaches in future work.

Future work could further explore adaptive threshold selection, 
online centroid updating as new clusters are formed, 
and incorporating citation graph or document-level context for improved disambiguation.

\section{Bibliographical References}\label{sec:reference}

\bibliographystyle{lrec2026-natbib}
\bibliography{somd2026}

\bibliographystylelanguageresource{lrec2026-natbib}

\end{document}